 \documentclass[pmlr,twocolumn,10pt]{jmlr} 

\usepackage{xcolor}
\definecolor{genorange}{RGB}{255,163,0}   
\definecolor{evalblue}{RGB}{0,176,240}    
\definecolor{refgreen}{RGB}{0,176,80}     

\usepackage{fontawesome} 

\usepackage{booktabs}
\usepackage{colortbl}   
\definecolor{rowgray}{gray}{0.92}
\usepackage{siunitx}
\usepackage{multirow}
\usepackage[switch]{lineno}

\theorembodyfont{\upshape}
\theoremheaderfont{\scshape}
\theorempostheader{:}
\theoremsep{\newline}

\jmlrvolume{XXX}
\jmlryear{2025}
\jmlrworkshop{Machine Learning for Health (ML4H) 2025} 

 \title[SFT-TA: Supervised Fine-Tuned Agents for Thematic Analysis]{SFT-TA: Supervised Fine-Tuned Agents in Multi-Agent LLMs \\ for Automated Inductive Thematic Analysis}

\author{%
\Name{Seungjun Yi} \Email{\textsc{charlie.yi@utexas.edu}}\\
\addr Department of Biomedical Engineering, University of Texas at Austin, Austin, TX, USA
\AND
\Name{Joakim Nguyen} \Email{\textsc{jhn001@utexas.edu}}\\
\addr Department of Computer Science, University of Texas at Austin, Austin, TX, USA
\AND
\Name{Huimin Xu} \Email{\textsc{huiminxu@utexas.edu}}\\
\addr School of Information, University of Texas at Austin, Austin, TX, USA
\AND
\Name{Terence Lim} \Email{\textsc{terence.lim@utexas.edu}}\\
\addr College of Natural Sciences, University of Texas at Austin, Austin, TX, USA \\ Graphen, Inc., New York, NY, USA
\AND
\Name{Joseph Skrovan} \Email{\textsc{jskrovan@fastmail.fm}}\\
\addr School of Information, University of Texas at Austin, Austin, TX, USA
\AND
\Name{Mehak Beri} \Email{\textsc{mb72384@eid.utexas.edu}}\\
\addr School of Information, University of Texas at Austin, Austin, TX, USA
\AND
\Name{Hitakshi Modi} \Email{\textsc{hitakshi.modi@austin.utexas.edu}}\\
\addr Department of Pediatrics, University of Texas at Austin, Austin, TX, USA
\AND
\Name{Andrew Well} \Email{\textsc{andrew.well@vumc.org}}\\
\addr Department of Cardiac Surgery, Division of Pediatric Cardiac Surgery,\\
Vanderbilt University School of Medicine, Nashville, TN, USA\\
Pediatric Heart Institute, Monroe Carell Jr. Children’s Hospital at Vanderbilt, Nashville, TN, USA
\AND
\Name{Liu Leqi} \Email{\textsc{leqi.liu@mccombs.utexas.edu}}\\
\addr Department of Information, Risk, and Operations Management, McCombs School of Business,\\
University of Texas at Austin, Austin, TX, USA\\
Machine Learning Laboratory, University of Texas at Austin, Austin, TX, USA\\
\AND
\Name{Mia Markey} \Email{\textsc{mia.markey@utexas.edu}}\\
\addr Department of Biomedical Engineering, University of Texas at Austin, Austin, TX, USA
\AND
\Name{Ying Ding} \Email{\textsc{ying.ding@ischool.utexas.edu}}\\
\addr
Bill \& Lewis Suit Professor, School of Information, University of Texas at Austin, Austin, TX, USA\\
Department of Population Health, Dell Medical School, University of Texas at Austin, Austin, TX, USA
}

\begin{document}

\maketitle

\begin{abstract}
Thematic Analysis (TA) is a widely used qualitative method that provides a structured yet flexible framework for identifying and reporting patterns in clinical interview transcripts.
However, manual thematic analysis is time-consuming and limits scalability.
Recent advances in LLMs offer a pathway to automate thematic analysis, but alignment with human results remains limited.
To address these limitations, we propose SFT-TA, an automated thematic analysis framework that embeds supervised fine-tuned (SFT) agents within a multi-agent system.
Our framework outperforms existing frameworks and the 
gpt-4o baseline in alignment with human reference themes.
We observed that SFT agents alone may underperform, but achieve better results than the baseline when embedded within a multi-agent system.
Our results highlight that embedding SFT agents in specific roles within a multi-agent system is a promising pathway to improve alignment with desired outputs for thematic analysis.


\end{abstract}



\paragraph*{Data and Code Availability}
The dataset consists of de-identified interview transcripts from nine focus groups with 42 parents of children diagnosed with anomalous aortic origin of a coronary artery (AAOCA). Sessions were lightly moderated to elicit open-ended discussions about parents’ lived experiences. The transcripts, collected between August 2021 and May 2022, had an average length of 10,987 words. The dataset originates from~\cite{mery2023journey}, and is not publicly available due to privacy concerns. However, it may be made available upon reasonable request. The code is not publicly available due to dataset privacy restrictions but may be shared upon reasonable request.


\paragraph*{Institutional Review Board (IRB)}
Data collection for the dataset used in this study was approved by the Institutional Review Board (IRB\# 2019080031). The study is registered at \href{https://clinicaltrials.gov/}{ClinicalTrials.gov }(NCT04613934).

\section{Introduction}
\label{sec:intro}

Thematic Analysis (TA) is one of the most widely used methods for qualitative data analysis, offering a structured yet flexible framework for identifying, analyzing, and reporting patterns within interview transcripts. 
Since its formalization by~\cite{braun2006thematic},
inductive thematic analysis has been the predominant method for analyzing interview transcripts to uncover actionable insights that inform decision-making across research and applied fields such as education, transportation, and customer experience.
Manual thematic analysis begins with researchers immersing themselves in the data through repeated readings and noting initial ideas (Steps 1–2). From these, systematic codes are generated (Step 3) to capture salient features of the data, which then serve as the foundation for searching for candidate themes (Step 4). These themes are subsequently reviewed, refined, and clearly defined and named (Step 5), before producing the final report in the form of a structured thematic framework that conveys the relationships and underlying meanings (Step 6).
However, this process remains highly resource-intensive and inefficient, consuming an estimated 6.1 million hours annually in the U.S. alone and posing a significant barrier to scalability and timely insight generation~\citep{campbell2021reflexive}.
Automating thematic analysis and reducing manual workload are essential for timely, scalable, and reproducible theme generation.

\begin{figure*}[t!]
    \centering
    \includegraphics[width=\textwidth]{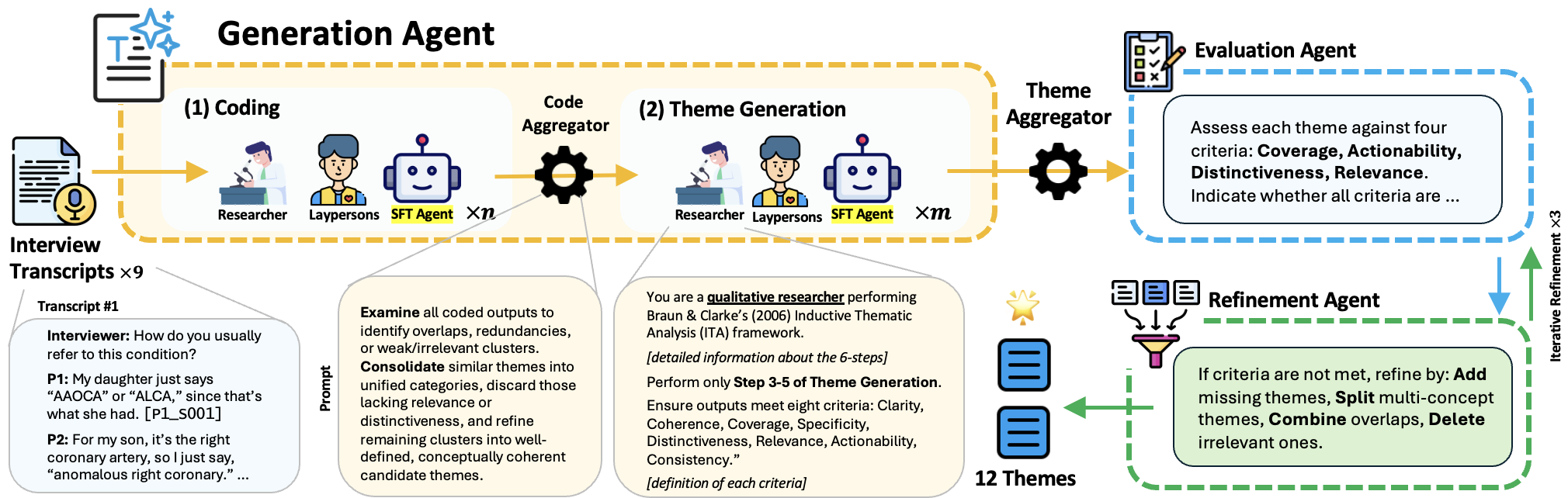}
    \caption{\textbf{Overview of the SFT-TA Framework.} Clinical focus group interview transcripts are processed through \textcolor{genorange}{coding}, \textcolor{genorange}{aggregation}, and \textcolor{genorange}{theme generation}, with the key novelty of embedding \textbf{SFT agents} alongside other identites (e.g., researcher, laypersons) in the generation steps. This is followed by three rounds of iterative refinement with the \textcolor{evalblue}{evaluation} and \textcolor{refgreen}{refinement} agents, ultimately yielding 12 themes. Each arrow represents a transformation from input $a \rightarrow b$, reflecting the role of the LLM and guided by a prompting template. \textcolor{genorange}{Orange arrows} denote \textbf{generation agents}, \textcolor{evalblue}{blue arrows} the \textbf{evaluation agent}, and \textcolor{refgreen}{green arrows} the \textbf{refinement agent}. Labels above the arrows indicate the agent responsible for each transformation.}
    \label{fig:placeholder}
\end{figure*}

Recent advances in large language models (LLMs) offer a promising alternative for automating key components of the six-step inductive thematic analysis process. LLMs can generate initial codes (Step 2)~\citep{yi2025protomedllmautomaticevaluationframework, yi2025autota} and candidate themes (Steps 3–5)~\citep{xu2025tamahumanaicollaborativethematic, raza2025llmtallmenhancedthematicanalysis} from interview transcripts within minutes, representing a substantial reduction compared to the several hours typically required by human analysts.
However, the alignment between generated results and manual thematic analysis (human reference themes) remains limited.

To address these limitations, we introduce \textbf{SFT-TA} (Figure~\ref{fig:placeholder}), an automated thematic analysis framework inspired by the collaborative nature of manual triangulation~\citep{braun2006thematic}. In qualitative research, \textit{triangulation} refers to the incorporation of multiple perspectives to interpret the same dataset, thereby ensuring that no single viewpoint dominates the analysis. Motivated by this principle, we posed the question:\textit{ Could introducing an agent with prior knowledge of the dataset and the analytic process improve alignment with human-generated themes?} By analogy, SFT-TA embeds supervised fine-tuned (SFT) agents into the generation step within a multi-agent system, enabling diverse “voices” within the system to collaboratively analyze transcripts and reduce bias of individual models.




In summary, our main contributions can be summarized as:

\begin{itemize}
    \item \textbf{Introducing SFT Agents into Multi-Agent Systems.} We introduce SFT-TA, a framework that embeds supervised fine-tuned (SFT) agents into multi-agent systems. In particular, we incorporate SFT agents into the generation phase, specifically within the coding and theme  generation steps.
    \item \textbf{Substantial Performance Gains.} We propose SFT-TA, a novel framework that integrates supervised fine-tuned agents into multi-agent systems, achieving significant improvements over the \texttt{gpt-4o} baseline: Fuzzy Match +10.3\%, Cosine Similarity +15.3\%, and METEOR +22.8\%.
    \item \textbf{Collaboration Transforms SFT Agent Performance.} We discover a notable phenomenon: while SFT agents alone may underperform relative to the baseline (e.g., Fuzzy Match --1.2\%, $\mathcal{C}$ --7.6\%), their integration within our multi-agent system yields substantial improvements, achieving +10.3\% in Fuzzy Match and +22.6\% in $\mathcal{C}$.
    \item \textbf{Optimal Dataset Size for SFT.}  
    We identify a "golden point" in dataset size for SFT agents in thematic analysis, at which performance saturates, consistent with prior findings~\citep{parfenova-etal-2025-text}.
\end{itemize}

\section{Related Work}

\paragraph{Automating Thematic Analysis (TA) with LLMs}

Recent efforts have explored the use of LLMs to automate various components of inductive thematic analysis (ITA), motivated by the need to reduce the manual workload involved in the process~\citep{yi2025positionthematicanalysisunstructured}. Most existing frameworks target specific stages of the ITA workflow: typically coding~\citep{prescott2024comparing, katz2024thematicanalysisopensourcegenerative}  or theme generation~\citep{Mannstadt2024Novel, deiner2024llmita}. Additionally, most approaches rely on single-agent LLMs without role specialization or supervision. While a few hybrid frameworks integrate human oversight~\citep{Dai2023LLMInLoop, xu2025tamahumanaicollaborativethematic}, they often require manual intervention at multiple stages, constraining scalability. Notably, only a small number of studies support all three core phases of ITA: initial coding, theme generation, and theme refinement or validation~\citep{raza2025llmtallmenhancedthematicanalysis, depaoli2023inductive, yi2025autota}. 




\paragraph{Multi-Agent Systems for Thematic Analysis}
Recent advances in thematic analysis increasingly leverage multi-agent systems integrated with llms to enhance scalability, diversity, and methodological rigor in processing large datasets. For example, \textit{Thematic-LM} \citep{qiao2025thematiclm} presents a multi-agent architecture where specialized coder agents, each embodying distinct identity perspectives, collaborate to code, aggregate, and update a shared codebook across large corpora data, thereby fostering divergent thematic interpretations and addressing limitations of sequential and human-biased analyses. Complementing this, \textit{TAMA} \citep{xu2025tamahumanaicollaborativethematic} offers a human-AI collaborative framework tailored to clinical interviews: multi-agent LLMs generate and refine themes under expert oversight, yielding superior thematic hit rate, coverage, and distinctiveness compared to single-agent approaches. Following the work, \textit{Auto-TA} \citep{yi2025autota} advances toward fully automated analysis by deploying multi-agent LLMs augmented with reinforcement learning from human feedback. Earlier,  \citep{rasheed2024llm} demonstrated that functional diversity among agents, such as summarizers, coders, and categorizers, can enrich interpretive depth in qualitative coding. Together, these systems demonstrate a clear progression from human-in-the-loop to autonomous multi-agent  implementations, each tackling challenges of scalability, coder bias, and codebook management.

\paragraph{SFT-TA: Supervised Fine-tuning for Multi-Agents}
Supervised fine-tuning (SFT) has emerged as a key technique to enhance the performance and coordination of multi-agent systems, particularly those integrating LLMs for complex tasks such as thematic analysis. In SFT, each agent is trained on labeled data to optimize its behavior or decision-making within a team, ensuring consistent adherence to task-specific standards while preserving inter-agent diversity \citep{brown2020language,ouyang2022training}. This approach has been applied in multi-agent systems for collaborative coding, where individual agents learn from human-annotated examples to produce more accurate, coherent, and reliable thematic outputs \citep{qiao2025thematiclm}. By fine-tuning agents in a supervised manner, systems can achieve better alignment with human interpretive practices, reduce coding errors, and maintain complementary behaviors across agents, which is crucial for mitigating redundancy and bias in multi-agent analyses \citep{xu2025tamahumanaicollaborativethematic,yi2025autota}. 
SFT accelerates convergence and provides a foundation for reinforcement learning or human-in-the-loop refinement, enabling multi-agent systems to scale to large, heterogeneous datasets while preserving interpretive depth and rigor \citep{drapal2023llmsupport,Dai2023LLMInLoop}.

\section{Method}

\subsection{The SFT-TA Framework.}

We introduce SFT-TA, an automated inductive thematic analysis framework that embeds supervised fine-tuned (SFT) agents in the generation step (Figure~\ref{fig:placeholder}). The pipeline mimics the six-phase framework of \cite{braun2006thematic}, with a focus on the coding and theme-generation phases. Specifically, interview transcripts are processed through two sub-steps: (1) \textit{Coding}, corresponding to Step~2 (\textit{generating initial codes}), and (2) \textit{Theme Generation}, corresponding to Steps~3--5 (\textit{searching for themes, reviewing themes, and defining/naming themes}). 

In the coding phase, the full dataset was distributed to multiple coder agents, each of which independently processed all transcripts.
As each of the nine transcripts represented a distinct context with different participants, they were also processed independently during the coding stage.
Their outputs were then consolidated by a coder aggregator. 
The aggregated codes are then passed to multiple theme-generation agents, which also operate independently across agents, and their outputs are consolidated by a theme aggregator to produce the final twelve set of themes. 
Among the multiple coder agents and theme-generation agents, a single SFT agent was incorporated at each stage (one within the coding phase and one within the theme-generation phase).
After generating twelve themes, the framework undergoes three iterative refinement rounds, guided by an evaluation agent and a refinement agent. During this process, limited human oversight is applied: themes deemed entirely off-topic are removed, and in practice some themes may undergo more or fewer than three refinement iterations.


\subsection{Dataset}
\label{sec:dataset}
This study uses a subset of interview transcripts described in the \textit{Data and Code Availability} section. For analysis, the transcripts were manually condensed to reduce token costs while preserving semantic content. Each quote was indexed with a unique Quote ID to support reproducible evaluation and consistent referencing in LLM-generated outputs. To benchmark performance, we use twelve human-generated themes from \citet{mery2023journey} as reference annotations. These human-generated themes, produced via inductive thematic analysis, serve as the ground truth against which we compare LLM-generated themes, hereafter referred to as the \textit{human reference themes}.
The complete list of twelve human reference themes is provided in Table~\ref{tab:theme_alignment}.


\subsection{Supervised Fine-Tuned (SFT) Agents and Dataset Augmentation}  
We performed supervised fine-tuning of \texttt{gpt-4o} via the \href{https://platform.openai.com/docs/guides/supervised-fine-tuning}{OpenAI API} using a pairwise dataset molded into a single \texttt{.jsonl} format. Each datapoint consisted of a prompt–answer pair, where the prompt was designed to perform inductive thematic analysis and generate themes from the original transcripts (see~\ref{sec:imp-details} for details), and the answer was a set of human-generated themes. In total, twelve human reference themes served as examples of desired output, with ten used for training and two reserved for validation. The ten themes were each paraphrased into thirty variants using multiple LLMs\footnote{\texttt{gpt-4o}, \texttt{gpt-o4-mini-high}, \texttt{gemini-2.5-flash}, and \texttt{TextGrad}~\citep{yuksekgonul2024textgradautomaticdifferentiationtext}} and reviewed by human annotators; based on this review, the \texttt{gpt-4o} paraphrased versions were selected for inclusion in the final fine-tuning dataset. An example of a paraphrased theme is provided in Appendix~\ref{app:theme-ex}.


Following \citet{parfenova-etal-2025-text}, which reports that fine-tuning performance saturates around 200-300 datapoints across diverse LLMs, we adopt a similar scale to mitigate annotation burdens—a key advantage in clinical domains where large, high-quality datasets are scarce. 
To validate this choice, we conducted an ablation study on SFT dataset size, which showed that performance peaked at approximately 300 datapoints.
Full results are provided in Appendix~\ref{app:dataset-ablation}.

\subsection{Implementation Details.}
\label{sec:imp-details}

\paragraph{Prompt Optimization.}

We incorporated the following prompt optimization strategies
to derive the maximum gains
prompt engineering itself.
Our main strategy is to replicate the six-phase procedures outlined in~\cite{braun2006thematic} through Chain-of-Thought (CoT) reasoning, explicitly specify the evaluation criteria defined by the cardiac surgeon as the desired output, and employ a a structured prompt format (e.g., using tags such as ``\texttt{< >}``) to increase the likelihood that LLMs will generate the intended outcomes.
At the final step, the model was explicitly prompted to generate exactly 12 themes.
In addition, a cardiac surgeon provided a set of eight criteria for coders to consider during the inductive thematic analysis process, consistent with those empirically applied in prior work~\cite{mery2023journey}. These criteria were incorporated into the prompting procedure, and LLMs were explicitly instructed that each generation step would be evaluated against them. The full list of criteria is presented in Appendix~\ref{app:eval-list}.
Also, we incorporated role identities in our prompts. For all generation tasks, the model was assigned the role of a qualitative researcher, reflecting the nature of inductive thematic analysis. Specifically, it was instructed to act as an expert qualitative researcher trained in the inductive thematic analysis framework of ~\cite{braun2006thematic}.


\subsection{Experiments}
We conducted an ablation study to evaluate the impact of incorporating SFT agents.  
A total of ten experimental conditions were tested, each corresponding to each row in Table~\ref{tab:results} and Table~\ref{tab:expert-eval}. Here, "\faAndroid" denotes a SFT agent. 

\begin{itemize}
    \item \textbf{Baselines:} (i) single-agent \texttt{gpt-4o}, (ii) single-agent \texttt{gpt-5}, and existing automated thematic analysis frameworks (iii) ~\cite{xu2025tamahumanaicollaborativethematic} and (iv) ~\cite{raza2025llmtallmenhancedthematicanalysis} all without SFT agents.
    \item \textbf{Vanilla (only \faAndroid):} SFT agents incorporated in both coding and theme-generation stages.
    \item \textbf{Frameworks w/ \faAndroid:}~\cite{xu2025tamahumanaicollaborativethematic} and~\cite{raza2025llmtallmenhancedthematicanalysis} with SFT agents added in the theme-generation stage.
    \item \textbf{Selective SFT Placement:} SFT agents incorporated exclusively in (i) coding or (ii) theme generation.
\end{itemize}



\subsection{Evaluation}
\label{sec:evaluation}

We evaluate the alignment between LLM-generated themes and human reference themes (see Section~\ref{sec:dataset} for details) using both traditional metrics and thematic analysis (TA)-specific metrics. For traditional metrics, we report both lexical-overlap and semantic-similarity measures. 
Formally, let $\mathcal{H} = \{h_1, \dots, h_{12}\}$ denote the 12 human reference themes and $\mathcal{G} = \{g_1, \dots, g_{12}\}$ denote the 12 LLM-generated themes. For a given evaluation metric $M(\cdot,\cdot)$, we compute $M(h_i, g_j)$ for every pair $(h_i, g_j)$, yielding $12 \times 12 = 144$ values in total. For each reference theme $h_i$, we take the maximum score across all generated themes, $\max_{j} M(h_i, g_j)$, and then average across all reference themes:
\[
S = \frac{1}{12} \sum_{i=1}^{12} \max_{j} M(h_i, g_j).
\]
Thus, each reference theme is aligned with its best-matching generated theme according to the chosen metric, and the final score $S$ is the average over all 12 reference themes.  
The detailed definitions of each evaluation metric are provided below.


\paragraph{Traditional Metrics.}

Traditional metrics can be grouped into two categories: (i) lexical-overlap metrics and (ii) semantic-similarity metrics.
(i) Lexical-overlap metrics evaluate surface-level similarity. Fuzzy Match captures string-level overlap using edit-distance–based alignment, while BLEU~\citep{papineni2002bleu}, METEOR~\citep{banerjee-lavie-2005-meteor}, and ROUGE-L~\citep{lin2004rouge} provide widely adopted measures of n-gram overlap.
(ii) Semantic-similarity metrics assess meanings beyond surface form. Cosine Similarity measures closeness in vector space using sentence embeddings.
All metrics yield scores in the range [0,1], with higher values indicating greater similarity. Full mathematical definitions and implementation details are provided in Appendix~\ref{app:metrics-details}.
These metrics complement each other but remain limited in capturing the conceptual focus of inductive thematic analysis beyond surface form. Therefore, we introduce additional metrics tailored specifically for evaluating thematic analysis, as described below.

\begin{table*}[t]
\centering
\caption{\textbf{Evaluation results on traditional and TA metrics.} Best values are in bold. \faAndroid\ denotes the SFT Agent. Delta values are relative to the Vanilla baseline, highlighted in gray. Results for the full SFT-TA pipeline are highlighted in yellow. Deltas in subscripts are computed relative to the \texttt{gpt-4o} baseline. \textbf{Bolded} values indicate the largest positive improvement in each column.}
\label{tab:results}
\scriptsize
\begin{tabular}{l cccc ccc}
\toprule
\multirow{2}{*}{\bfseries Method}
& \multicolumn{4}{c}{\bfseries Traditional Metrics}
& \multicolumn{3}{c}{\bfseries TA Metrics} \\
\cmidrule(lr){2-5} \cmidrule(lr){6-8}
& Fuzzy & Cosine & BLEU & METEOR & $\mathcal{C}$ & $\mathcal{D}$ & $\mathcal{T}$ \\
\midrule
\multicolumn{8}{l}{\textit{Baseline (w/o \faAndroid)}} \\
\midrule
\rowcolor{gray!15} Vanilla (\texttt{gpt-4o}) 
& .457 & .115 & .058 & .120 & .557$_{(.000)}$ & .407$_{(.000)}$ & .308$_{(.000)}$ \\
Vanilla (\texttt{gpt-5}) 
& .416$_{(-.041)}$ & .078$_{(-.037)}$ & .068$_{(+.010)}$ & .084$_{(-.036)}$ & .439$_{(-.118)}$ & .273$_{(-.134)}$ & .293$_{(-.015)}$ \\
\cite{xu2025tamahumanaicollaborativethematic} 
& .495$_{(+.038)}$ & .144$_{(+.029)}$ & .040$_{(-.018)}$ & .144$_{(+.024)}$ & .592$_{(+.035)}$ & .191$_{(-.216)}$ & .251$_{(-.057)}$ \\
\cite{raza2025llmtallmenhancedthematicanalysis} 
& .497$_{(+.040)}$ & .167$_{(+.052)}$ & .040$_{(-.018)}$ & .184$_{(+.064)}$ & .635$_{(+.078)}$ & .176$_{(-.231)}$ & .301$_{(-.007)}$ \\
\midrule
\multicolumn{8}{l}{\textit{Fine-tuned (w/ \faAndroid)}} \\
\midrule
Vanilla (only \faAndroid) 
& .445$_{(-.012)}$ & .124$_{(+.009)}$ & .072$_{(+.014)}$ & .179$_{(+.059)}$ & .481$_{(-.076)}$ & .104$_{(-.303)}$ & .281$_{(-.027)}$ \\
\cite{xu2025tamahumanaicollaborativethematic} w/ \faAndroid
& .453$_{(-.004)}$ & .143$_{(+.028)}$ & .058$_{(.000)}$ & .187$_{(+.067)}$ & .503$_{(-.054)}$ & .214$_{(-.193)}$ & \textbf{.315}$_{(+.007)}$ \\
\cite{raza2025llmtallmenhancedthematicanalysis} w/ \faAndroid 
& .483$_{(+.026)}$ & .161$_{(+.046)}$ & .056$_{(-.002)}$ & .276$_{(+.156)}$ & .636$_{(+.079)}$ & .182$_{(-.225)}$ & .265$_{(-.043)}$ \\
\faAndroid \space in coding only 
& .518$_{(+.061)}$ & .225$_{(+.110)}$ & .072$_{(+.014)}$ & .274$_{(+.154)}$ & .752$_{(+.195)}$ & .217$_{(-.190)}$ & .297$_{(-.011)}$ \\
\faAndroid \space in TG only 
& .530$_{(+.073)}$ & .254$_{(+.139)}$ & \textbf{.094}$_{(+.036)}$ & .338$_{(+.218)}$ & \textbf{.787}$_{(+.230)}$ & .322$_{(-.085)}$ & .292$_{(-.016)}$ \\
\rowcolor{yellow!30} \faAndroid \space in Coding \& TG
& \textbf{.560}$_{(+.103)}$ & \textbf{.268}$_{(+.153)}$ & .080$_{(+.022)}$ & \textbf{.348}$_{(+.228)}$ & .783$_{(+.226)}$ & \textbf{.394$_{(-.013)}$} & .303$_{(-.005)}$ \\
\bottomrule
\end{tabular}
\end{table*}

\paragraph{Metrics Specific to Inductive Thematic Analysis: $\mathcal{C}$, $\mathcal{D}$, $\mathcal{T}$.}

We adopt the methodology of previous work~\cite{qiao2025thematiclm}, which applied the four trustworthiness principles commonly used to evaluate thematic analysis (TA) results from~\cite{Stahl2020UnderstandingTrustworthiness}. These principles include credibility, confirmability, dependability, and transferability, and were consolidated into the following three evaluation metrics. 

\paragraph{Credibility and Confirmability ($\mathcal{C}$)} measures whether the generated themes accurately represent the underlying data rather than being influenced by biases. Each quote in the transcript is assigned a unique identifier, referred to as a Quote ID (e.g., \texttt{[P1\_S002]} for Participant 1, Sentence 2). We link each theme to its associated data segments through these quote IDs and assign an evaluator agent to calculate the percentage of cases where the quotes and themes are consistent. Inconsistencies may arise from hallucinations or internal biases within the LLM.

\begin{table*}[t]
\centering
\caption{\textbf{Human Evaluation.} Comparison across coverage, actionability, distinctiveness, and relevance using a 1--5 Likert scale, where 1 = Completely disagree, 2 = Disagree, 3 = Neither agree nor disagree, 4 = Agree, and 5 = Completely agree (relative to the statement in Appendix~\ref{appendix:human_eval}). TG denotes the theme-generation stage.
Deltas (in subscripts) are computed relative to the baseline \texttt{gpt-4o}. \textbf{Bolded} values indicate the largest positive improvement in each column.}
\label{tab:expert-eval}
\resizebox{\linewidth}{!}{%
\begin{tabular}{lcccc}
\hline
\textbf{Methods} & \textbf{Coverage} & \textbf{Actionability} & \textbf{Distinctiveness} & \textbf{Relevance} \\
\hline
\multicolumn{5}{l}{\textit{Baseline (w/o \faAndroid)}} \\
\hline
\rowcolor{gray!15} Vanilla (\texttt{gpt-4o}) & 4.00 & 2.80 & 3.58 & 4.38 \\
Vanilla (\texttt{gpt-5}) & 2.50\textsubscript{(-1.50)} & 3.79\textsubscript{(+0.99)} & 3.65\textsubscript{(+0.07)} & 3.74\textsubscript{(-0.64)} \\
\cite{xu2025tamahumanaicollaborativethematic} & 3.50\textsubscript{(-0.50)} & 3.50\textsubscript{(+0.70)} & 4.50\textsubscript{(+0.92)} & 4.34\textsubscript{(-0.04)} \\
\cite{raza2025llmtallmenhancedthematicanalysis} & 4.50\textsubscript{(+0.50)} & 3.61\textsubscript{(+0.81)} & 4.66\textsubscript{(+1.08)} & 4.30\textsubscript{(-0.08)} \\
\hline
\multicolumn{5}{l}{\textit{Fine-tuned (w/ \faAndroid)}} \\
\hline
Vanilla (only \faAndroid) & 4.00\textsubscript{(0.00)} & 3.47\textsubscript{(+0.67)} & 4.82\textsubscript{(+1.24)} & 4.00\textsubscript{(-0.38)} \\
\cite{xu2025tamahumanaicollaborativethematic} w/ \faAndroid & 4.00\textsubscript{(0.00)} & 3.53\textsubscript{(+0.73)} & \textbf{4.86}\textsubscript{\textbf{(+1.28)}} & 3.97\textsubscript{(-0.41)} \\
\cite{raza2025llmtallmenhancedthematicanalysis} w/ \faAndroid & 3.50\textsubscript{(-0.50)} & 3.31\textsubscript{(+0.51)} & 4.64\textsubscript{(+1.06)} & 4.05\textsubscript{(-0.33)} \\
\faAndroid \space in coding only & 4.00\textsubscript{(0.00)} & 3.41\textsubscript{(+0.61)} & 3.91\textsubscript{(+0.33)} & 4.50\textsubscript{(+0.12)} \\
\faAndroid \space in TG only & 3.50\textsubscript{(-0.50)} & 3.56\textsubscript{(+0.76)} & 3.94\textsubscript{(+0.36)} & 4.28\textsubscript{(-0.10)} \\
\rowcolor{yellow!30} \faAndroid \space in Coding \& TG & \textbf{5.00}\textsubscript{\textbf{(+1.00)}} & \textbf{4.28}\textsubscript{\textbf{(+1.48)}} & 4.35\textsubscript{(+0.77)} & \textbf{4.63}\textsubscript{\textbf{(+0.25)}} \\
\hline
\end{tabular}%
}
\end{table*}

\paragraph{Dependability ($\mathcal{D}$)}
measures the consistency of theme generation across multiple independent runs. To quantify this, we calculate bidirectional ROUGE scores~\citep{lin2004rouge} for both ROUGE-1 (n=1, unigrams) and ROUGE-2 (n=2, bigrams) overlaps. For two theme sets A and B produced from separate runs, the directional and averaged scores are defined as:
\begin{align*}
\mathcal{R}_n^{A \rightarrow B} &=
\frac{|\text{$n$-grams}(A) \cap \text{$n$-grams}(B)|}{|\text{$n$-grams}(A)|} \\
\mathcal{R}_n &= \frac{\mathcal{R}_n^{A \rightarrow B} + \mathcal{R}_n^{B \rightarrow A}}{2} \\
\mathcal{D} &= \frac{\mathcal{R}_1 + \mathcal{R}_2}{2}
\end{align*}
We compute $\mathcal{D}$ over five independent runs and report the mean across all runs. Higher $\mathcal{D}$ values correspond to greater stability between runs. We report the improvement ($\Delta \mathcal{D}$) over a baseline configuration where one agent performs coding and another generates themes.

\paragraph{Transferability ($\mathcal{T}$)}  
measures how well themes $\Theta$ generated from one dataset $D$ apply to a new but contextually similar dataset $D'$. In our framework, we assess $\mathcal{T}$ by splitting the corpus into a training set (7 transcripts) and a validation set (2 transcripts), producing $\Theta_{\text{train}}$ and $\Theta_{\text{val}}$ separately. The computation follows the same bidirectional ROUGE procedure as for $\mathcal{D}$, except here $A = \Theta_{\text{train}}$ and $B = \Theta_{\text{val}}$, with $\mathcal{R}_n = \mathcal{R}_n^{A \rightarrow B}$.  
We evaluate $\mathcal{T}$ over all $\binom{9}{2} = 36$ possible splits, reporting the mean across these runs. Higher $\mathcal{T}$ values indicate that themes generalize well between subsets, reflecting stronger transfer.

\paragraph{Human Evaluation.}
We evaluated the generated themes with four human raters: a cardiac surgeon, a qualitative analyst with extensive experience in inductive thematic analysis, and two independent laypersons. The laypersons had no professional qualitative analysis experience and were only provided general information about the dataset\footnote{Includes the main disease (AAOCA), the participant group (parents of affected children), and the purpose of the interviews.}. All raters received written definitions of the evaluation criteria but were not shown examples to avoid bias. The theme-generation method was blinded: raters were not informed about which model or baseline produced a given theme set, nor were they given any methodological details (e.g., prompts, number of agents, or training). Each set of themes was presented in a randomized order with only the definitions of the evaluation criteria provided.
Evaluators were instructed to assess the quality of themes according to four criteria, each rated on a 5-point Likert scale (1 = strongly disagree, 5 = strongly agree):
\begin{enumerate}
\item \textbf{Coverage.} Whether the 12 generated themes collectively captured the breadth of parents’ lived experiences. This was assessed only by raters familiar with the dataset, with one rating assigned per theme set.
\item \textbf{Actionability.} Whether each theme encapsulated a clear, specific, and meaningful insight that could inform interventions, resources, or future research.
\item \textbf{Distinctiveness.} Whether each theme was unique, avoiding overlaps or redundancies.
\item \textbf{Relevance.} Whether each theme accurately reflected parents’ concerns and needs, without conflating them with patient perspectives.
\end{enumerate}

The full guiding statements used for each criterion are provided in Appendix~\ref{appendix:human_eval}.
For Actionability, Distinctiveness, and Relevance, raters scored each theme individually; scores were then averaged across themes and raters to represent the set. Coverage was scored once per set of themes. The criteria were adapted from prior manual thematic analysis of the same dataset~\cite{mery2023journey} and align with established evaluation practices in thematic analysis~\citep{xu2025tamahumanaicollaborativethematic}.
The aggregated ratings, averaged across all raters, are reported in Table~\ref{tab:expert-eval}.

\begin{table*}[t]
\centering
\caption{\textbf{Human reference vs. SFT-TA generated themes.} Each SFT-TA theme is paired with its closest human theme based on cosine similarity, with minor manual reordering applied.}
\label{tab:theme_alignment}
\renewcommand{\arraystretch}{1.25}
\scriptsize
\begin{tabular}{@{}p{0.48\textwidth} p{0.48\textwidth}@{}}
\toprule
\textbf{Human Reference Themes} & \textbf{SFT-TA Themes} \\
\midrule
Worrying too much lessened when risks are explained & Clarity of potential risks and outcomes \\
Diagnosis told with care and kindness & The diagnosis given in a compassionate and empathic way \\
Families want steady information that makes sense & Being appropriately informed \\
Feeling in control of decisions about future & A sense of control over the future \\
Having trust and respect from doctors & Being heard and taken seriously by clinicians \\
Parents needing comfort and support from peers & Receiving support from others \\
Stress, anxiety, and sadness are still common & Appropriately coping with stress, anxiety and depression \\
Not blaming self for when or why diagnosis happened & Not feeling responsible for the diagnosis and its timing \\
Confidence that child is protected and safe & Feeling that my child is safe \\
Parents and clinicians working together as partners & Partnership with the care team \\
Guidance that fits family’s own choices & Individualized support for management decision-making \\
Support from faith or community when needed & Freedom from hypervigilance related to the condition \\
\bottomrule
\end{tabular}
\end{table*}



\section{Results}

\paragraph{Evaluation results on traditional and TA metrics.}
We report evaluation results in Table~\ref{tab:results}, while Table~\ref{tab:theme_alignment} presents the two generated themes corresponding to the final row of Tables~\ref{tab:results} and~\ref{tab:expert-eval}. Our framework achieved the best improvements across Fuzzy Match, Cosine Similarity, and METEOR, yielding performance gains of +10.3\%, +15.3\%, and +22.8\%, respectively (highlighted in yellow).
Compared to the vanilla \texttt{gpt-4o} baseline, the SFT agent alone did not improve performance and in some cases performed worse (e.g., $\mathcal{D}$ = 0.104). However, when integrated into the coding step, results exceeded prior frameworks~\citep{raza2025llmtallmenhancedthematicanalysis, xu2025tamahumanaicollaborativethematic}. Incorporation in the theme generation step further improved performance . Our SFT-TA framework with SFT agents in both coding and theme generation achieved the best outcomes across traditional metrics—Fuzzy = 0.560 (+0.103), Cosine = 0.268 (+0.153), METEOR = 0.348 (+0.228)—and also yielded the strongest dependability score $\mathcal{D}$ = 0.394. These results demonstrate that while fine-tuning alone is insufficient, embedding SFT agents into both coding and theme generation step in thematic analysis yields the largest overall improvements.

 For TA-specific metrics, we also observe that Dependability ($\mathcal{D}$) decreases relative to the baseline. For example, the Vanilla (\texttt{gpt-4o}) baseline achieves $\mathcal{D} = .407$, whereas the best-performing full SFT-TA pipeline (SFT in Coding \& TG) yields $\mathcal{D} = .394$, a slight decline ($-0.013$). This degradation may stem from additional complexity or stochasticity introduced by increasing the number of agents, which in turn reduces reproducibility. Future work should examine the underlying causes of this degradation, including whether tuning parameters of individual agents can mitigate its impact on $\mathcal{D}$. 

\paragraph{Human Evaluation Results.}

Table~\ref{tab:expert-eval} presents the results of the human evaluation. 
Our framework obtained the highest scores across coverage, actionability, and relevance, with Coverage = 5.00 (+1.00), Actionability = 4.28 (+1.48), and Relevance = 4.63 (+0.25), as shown in Table~\ref{tab:expert-eval} (highlighted in yellow).
However, we observed a degradation in distinctiveness (4.35, +0.77), the underlying cause of which remains unclear.
One possible explanation is that prior studies on multi-agent systems for thematic analysis~\cite{qiao2025thematiclm} have focused primarily on short-context datasets (e.g., Reddit threads), whereas our framework operates on substantially longer transcripts. In this setting, TA-specific metrics ($\mathcal{C}$, $\mathcal{D}$, $\mathcal{T}$) may be less well-suited for evaluating long-context data, such as the clinical transcripts used here. This remains a hypothesis, and further investigation is needed to better understand the observed phenomenon.

Incorporating SFT agents into prior frameworks, or limiting them to either the coding or theme-generation stage, resulted in lower performance than existing methods across coverage, actionability, and relevance. Moreover, our findings indicate that human evaluation results do not consistently align with traditional or TA-specific metrics.





\paragraph{Conclusion.}

We introduced a novel framework for automating inductive thematic analysis of clinical transcripts, centered on the integration of SFT agents within a multi-agent architecture. Our experiments across ten conditions demonstrate that while SFT agents alone may not outperform existing baselines, their incorporation into the full pipeline yields substantial improvements on traditional metrics, highlighting their value in complex qualitative analysis tasks. Although this study focused on clinical transcripts related to AAOCA, the framework generalizes to other domains and datasets, offering a scalable path toward LLM-assisted qualitative research.

\paragraph{Limitations and Future Work.}

The scope of this study is limited to nine clinical transcripts pertaining to AAOCA. Consequently, the transferability of the proposed framework to other datasets remains uncertain.
The human reference themes are not exhaustive. Although they were generated by domain experts following the procedures of inductive thematic analysis outlined in the original work \citep{braun2006thematic}, they should not be considered the definitive output of such an analysis. We also observed that some LLM-generated themes represent lower-level or more granular interpretations of the corresponding human-derived themes.
Future directions include integrating SFT agents into additional components of the pipeline—particularly evaluation and refinement agents—and conducting further ablation studies on the effect of varying the number of SFT agents.
For the baselines, future work should explore additional single-agent models beyond the \texttt{gpt} variants.
Another direction for future work is to explore LLM-as-a-judge approaches as a part of evaluation, such as Win Rate (see Appendix~\ref{app:win-rate}), to evaluate theme quality without relying solely on human ground truth. This would also enable investigation of the correlation between human judgments and these automated metrics.


\clearpage
\bibliography{jmlr-sample}

\appendix

\section{Criteria for Evaluating Generated Themes}\label{app:eval-list}

The following eight criteria, adapted from \citet{mery2023journey}, were used to assess the quality of generated themes:

\begin{enumerate}
    \item \textbf{Uniqueness} – Each theme represents a distinct idea.
    \item \textbf{Actionability} – Each theme can inform interventions, support strategies, or policy decisions.
    \item \textbf{Specificity} – Each theme is concrete and not overly general.
    \item \textbf{Condition Relevance} – The theme clearly relates to the target condition (i.e., AAOCA).
    \item \textbf{Segment Focus} – The theme captures the perspective of the specified participant group (e.g., patient, parent).
    \item \textbf{Clarity} – The theme is easily understandable.
    \item \textbf{Plausibility} – The theme is thematically consistent and reasonable for this topic.
    \item \textbf{Overall Quality} – The theme demonstrates high conceptual and linguistic quality.
\end{enumerate}

\section{Examples of Paraphrased Human Reference Themes}\label{app:theme-ex}

Original: \textit{Clarity of potential risks and outcomes}.  
Paraphrases: Clear understanding of possible dangers and results; Clear insight into foreseeable risks and implications; Transparency regarding expected risks and consequences; Awareness of potential hazards and their effects; Comprehension of likely threats and outcomes; Recognition of possible negative outcomes and associated risks; Knowledge of potential challenges and their consequences; Clear depiction of likely risks and results; Understanding the implications of possible risks; Clear communication about risks and expected consequences.



\section{Definition of Win Rate ($\mathcal{W}$)}\label{apd:second}
\label{app:win-rate}
\paragraph{Win Rate ($\mathcal{W}$)}  
is the proportion of evaluation instances in which the model's output is preferred over a baseline according to a specified evaluation criterion. In our setting, \texttt{gpt-4o} served as the evaluator, was explicitly informed that the task concerned the generation of themes, and received the outputs in a randomized order (i.e., A vs.\ B) to mitigate position bias. Formally, if $N_{\text{win}}$ denotes the number of instances where the model's output is preferred and $N_{\text{total}}$ is the total number of evaluated instances, then:
\begin{equation*}
\mathcal{W} = \frac{N_{\text{win}}}{N_{\text{total}}}.
\end{equation*}

To evaluate whether observed win rates differed from chance, we computed Wilson 95\% confidence intervals and conducted one-sided binomial tests against a 50\% null hypothesis, applying Holm–Bonferroni correction across conditions. With $N=100$ paired outputs, power analyses indicated that effects of roughly 20 points above chance are detectable at $\alpha=0.05$. Accordingly, we set a threshold of $>$0.70 as the criterion for declaring reliable performance gains over the human baseline. Full calculation details are provided in Appendix~\ref{apd:calculations}.

\section{Effect of Dataset Size on SFT Performance}\label{app:dataset-ablation}

\textbf{Data Scaling and Saturation.} Generation agents are fine-tuned with training subsets of size $N \in \{100, 200, 300, 600, 900\}$ to examine the relationship between dataset size and performance, and to identify the saturation point at which additional data yield diminishing returns.

Increasing the number of supervised fine-tuning examples led to steady improvements in win rate ($\mathcal{W}$), with performance reliably exceeding chance once training size surpassed a few hundred examples. The most notable gain occurred between 100 and 300 samples, after which improvements diminished, indicating diminishing returns from additional data. This trend suggests that approximately 300 labeled examples are sufficient to reach stable performance, while larger datasets contribute primarily to reducing variability across runs rather than to further accuracy gains.

\begin{figure}[htbp]
    \centering
    \includegraphics[width=0.50\textwidth]{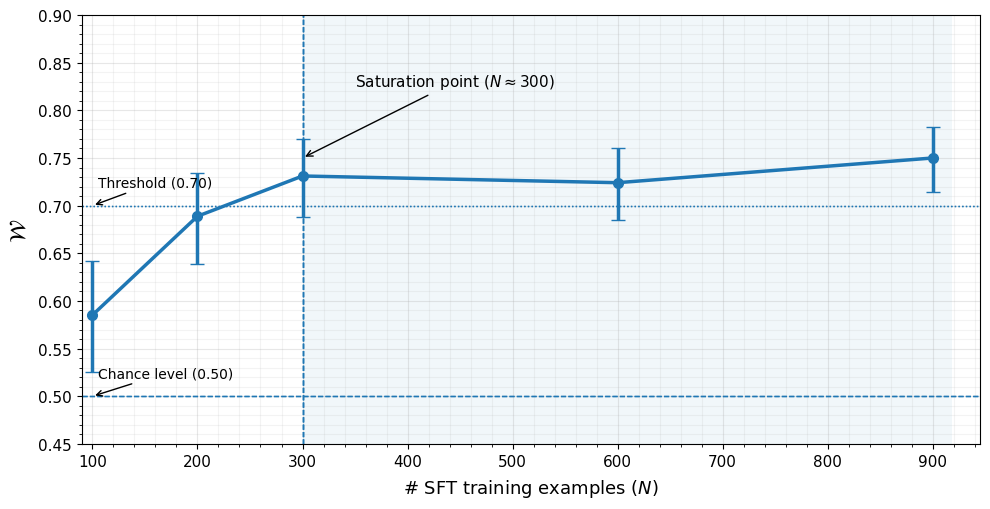}
    \caption{Win rate ($\mathcal{W}$) as a function of supervised fine-tuning (SFT) training examples ($N$). Dashed lines mark the chance level (0.50) and reliability threshold (0.70). A saturation point appears near $N \approx 300$. Error bars denote 95\% confidence intervals. The rationale for selecting 0.70 as the threshold is provided in Appendix~\ref{apd:calculations}.}
    \label{fig:sft_saturation}
\end{figure}

\section{Statistical Basis for Threshold Determination}\label{apd:calculations}

Observed win rates $\hat{p}$ based on $n$ paired comparisons were evaluated through a combination of confidence intervals, hypothesis testing, and power analysis. Confidence intervals were estimated using the Wilson score method rather than the normal approximation, as Wilson intervals are known to provide more accurate coverage probabilities for moderate sample sizes and probabilities near the boundaries of 0 and 1. For a two-sided 95\% interval, the formula is
\[
\hat{p}_\text{Wilson} = 
\frac{\hat{p} + \tfrac{z^2}{2n} \pm 
z \sqrt{\tfrac{\hat{p}(1-\hat{p})}{n} + \tfrac{z^2}{4n^2}}}{1 + \tfrac{z^2}{n}},
\quad z = 1.96.
\]
Intervals not containing 0.50 are interpreted as evidence that performance differs from chance at the 5\% level.

For formal hypothesis testing, we applied exact one-sided binomial tests to assess whether observed win counts exceeded chance. Specifically, if $x$ out of $n$ outputs favored the model, the $p$-value is
\[
p = \sum_{k=x}^{n} \binom{n}{k} (0.5)^n.
\]
This formulation directly evaluates the probability of observing at least $x$ wins under the null hypothesis that each trial has probability $0.5$. To account for testing across multiple experimental conditions, we applied Holm--Bonferroni correction, which sequentially adjusts significance thresholds while maintaining familywise error control.

To determine appropriate sample sizes, we conducted power analyses for the one-sided binomial test. Power calculations indicated that with $n=100$ paired outputs, the smallest reliably detectable effect size at $\alpha=0.05$ with 95\% power is approximately 0.20, corresponding to a difference between a 0.50 null rate and a 0.70 observed rate. Thus, effects smaller than 70\% vs.~50\% would be underpowered with this design, whereas effects at or above this level are detectable with high confidence. 

Based on these considerations, we adopted a threshold of $>0.70$ win rate as the criterion for declaring reliable performance gains relative to the human baseline. This choice integrates both statistical significance (through binomial testing with correction) and practical detectability (through power analysis), ensuring that reported improvements reflect robust evidence rather than chance fluctuations.

\section{Lexical Overlap Metrics Details}\label{app:metrics-details}

\paragraph{Fuzzy Match.}

We use fuzzy string matching to quantify the lexical similarity between LLM-generated and human-generated reference themes. Specifically, we leverage the \texttt{fuzzy\_match} metric implemented in \href{https://github.com/openai/evals}{OpenAI Evals}, which is built on normalized Levenshtein distance and token alignment heuristics. This allows for partial credit when the model output closely resembles the reference theme. The definition for mathematical notations is as follows.
Let \( s_1, s_2 \in \mathcal{S} \) be two strings. The fuzzy match score \( F(s_1, s_2) \in [0, 1] \) is computed as:

\[
F(s_1, s_2) = 1 - \frac{D_{\text{lev}}(s_1, s_2)}{\max(|s_1|, |s_2|)}
\]

where \( D_{\text{lev}} \) denotes the Levenshtein distance. A score of 1.0 implies an exact match, while lower scores reflect the degree of edit distance required. This metric is robust to minor lexical variations (e.g., \textit{"parental concern"} vs. \textit{"concerns of parents"}) and is particularly useful when evaluating generated themes that are semantically close but not identical in form. All evaluations were performed using \href{https://github.com/openai/evals}{OpenAI’s \texttt{fuzzy\_match} implementation}.

\paragraph{Cosine Similarity.}

We use cosine similarity to evaluate the semantic closeness between LLM-generated and human-generated reference themes based on their vector representations. Specifically, we embed each string using a pre-trained sentence encoder and compute the cosine of the angle between the resulting vectors. Let \( \vec{u}, \vec{v} \in \mathbb{R}^d \) be the embeddings of two theme strings. The cosine similarity score \( C(\vec{u}, \vec{v}) \in [-1, 1] \) is defined as:

\[
C(\vec{u}, \vec{v}) = \frac{\vec{u} \cdot \vec{v}}{\|\vec{u}\| \|\vec{v}\|}
\]

where \( \cdot \) denotes the dot product and \( \| \cdot \| \) the Euclidean norm. A score of 1.0 indicates perfect alignment, 0 implies orthogonality (no semantic overlap), and –1 reflects diametrically opposed meanings. We use encoder models such as \texttt{text-embedding-3-large}\footnote{Text embedding model from OpenAI, which supports up to 3072-dimensional embeddings and shows strong benchmark performance.} to obtain dense representations of text. 
This metric captures meaning beyond surface-level form and is especially valuable when comparing paraphrased or reworded themes that differ lexically but align semantically (e.g., \textit{``navigating complex medical decisions''} vs. \textit{``making difficult treatment choices''}, or \textit{``emotional response to diagnosis''} vs. \textit{``coping with the news''}).

\textbf{BLEU}~\citep{papineni2002bleu} computes the precision of overlapping n-grams, applying a brevity penalty to discourage short outputs. It tends to under-reward paraphrases or semantically aligned variants due to its strict reliance on exact matches. For example, \textit{``managing stress and anxiety''} vs. \textit{``handling emotional pressure''} would receive a low BLEU score despite similar meaning.

\textbf{METEOR}~\citep{banerjee-lavie-2005-meteor} addresses BLEU's limitations by incorporating unigram recall, stemming, synonym matching (via WordNet), and a fragmentation penalty. It can give partial credit to pairs like \textit{``healthcare access barriers''} and \textit{``difficulty reaching medical services''} even without exact word matches.

\textbf{ROUGE-L}~\citep{lin2004rouge} measures the longest common subsequence between reference and candidate texts, capturing both content and order overlap. For instance, \textit{``struggles after diagnosis''} and \textit{``challenges following diagnosis''} would yield moderate ROUGE-L scores due to shared sequential structure.

\section{Human Evaluation Guiding Statements}\label{appendix:human_eval}

For the human evaluation, raters were instructed to assess themes on a 5-point Likert scale  
(1 = strongly disagree, 5 = strongly agree) according to the following guiding statements:

\begin{itemize}
    \item \textbf{Coverage:} These themes comprehensively capture parents’ lived experiences and are supported by multiple participants.  
    \item \textbf{Actionability:} This theme provides clear, specific insights that can inform interventions, resources, or research.  
    \item \textbf{Distinctiveness:} This theme is clearly distinct from other themes, with no major overlaps or redundancies.  
    \item \textbf{Relevance:} This theme accurately reflects parents’ lived experiences and is directly related to having a child with AAOCA.  
\end{itemize}



\end{document}